\documentclass[letterpaper, 10 pt, conference]{IEEEtran} 

\usepackage{graphics} 
\usepackage{epsfig} 
\usepackage{amsmath} 
\usepackage{amssymb}  

\usepackage[utf8]{inputenc} 
\usepackage[T1]{fontenc}    

\usepackage[dvipsnames]{xcolor}
\usepackage{booktabs}       
\usepackage{amsfonts}       
\usepackage{nicefrac}       
\usepackage{microtype}      

\usepackage{booktabs} 
\usepackage{multirow}
\usepackage{pifont}
\usepackage{setspace}
\usepackage{comment}
\usepackage{algorithm}
\usepackage[noend]{algpseudocode}

\usepackage{verbatim}
\usepackage{caption}
\usepackage{wrapfig}
\usepackage{svg}
\usepackage{mathabx}
\usepackage{subcaption}
\usepackage{enumitem}
\usepackage{color,soul}
\usepackage{xfakebold}

\usepackage[noadjust]{cite}
\usepackage[english]{babel}
\usepackage{amsthm}
\theoremstyle{definition}

\theoremstyle{remark}

\setul{0.5ex}{0.3ex}
\definecolor{grey}{rgb}{.80,.80,0.80}
\setulcolor{grey}

\usepackage{hyperref}
\hypersetup{
    colorlinks=true,
    linkcolor=orange,
    filecolor=magenta,      
    urlcolor=orange,
    citecolor=orange,
}

\IEEEoverridecommandlockouts                              

\usepackage{amsfonts}               
\usepackage{amsmath}                
\usepackage{booktabs}               
\usepackage[T1]{fontenc}            
\usepackage{graphicx}               
\usepackage{hyperref}               
\usepackage{inconsolata}            
\usepackage[numbers,sort&compress]{natbib}        
\usepackage{nicefrac}               
\usepackage{xcolor}                 
\usepackage{multicol}
\usepackage[frozencache,cachedir=.]{minted}

\usepackage[utf8]{inputenc} 
\usepackage[T1]{fontenc}    
\usepackage{hyperref}       
\usepackage{url}            
\usepackage{booktabs}       
\usepackage{amsfonts}       
\usepackage{nicefrac}       
\usepackage{microtype}      
\usepackage{xcolor}         

\hypersetup{
    colorlinks=true,
    linkcolor=orange,
    filecolor=magenta,      
    urlcolor=orange,
    citecolor=orange,
}

\usepackage{titlesec}
\usepackage{xspace}
\usepackage{mathtools}
\usepackage{bbm}
\usepackage{enumitem}
\usepackage{caption}
\usepackage{subcaption}
\usepackage{pdflscape}
\usepackage{xcolor}
\usepackage[]{minted}
\setcitestyle{numbers,square,comma}
\usepackage{amsthm}
\usepackage{amsmath}

\usepackage[capitalise, nameinlink]{cleveref}

\newcommand{\fbseries}{\unskip\setBold\aftergroup\unsetBold\aftergroup\ignorespaces}
\makeatletter
\newcommand{\setBoldness}[1]{\def\fake@bold{#1}}
\newcommand{\ssymbol}[1]{^{\@fnsymbol{#1}}}
\makeatother

\newcommand{\ACRO}{{DROC}\xspace}

\newcommand{\eg}{e.g., }

\title{\Large \bf
Distilling and Retrieving Generalizable Knowledge for \\ Robot Manipulation via Language Corrections
}

\author{
Lihan Zha$\ssymbol{3}$, Yuchen Cui$\ssymbol{3}$, Li-Heng Lin$\ssymbol{3}$, Minae Kwon$\ssymbol{3}$, Montserrat Gonzalez Arenas$\ssymbol{4}$,\\
Andy Zeng$\ssymbol{4}$, Fei Xia$\ssymbol{4}$, Dorsa Sadigh$\ssymbol{3}$
\thanks{ 
$\ssymbol{3}$ Computer Science Department, Stanford University, Stanford, CA, USA. \newline
$\ssymbol{4}$ Google Deepmind, Moutain View, CA. \newline
Corresponding Email: {{\tt\footnotesize yuchenc@stanford.edu}}
}
\vspace{-0.2cm}
}

\begin{document}

\maketitle
\thispagestyle{empty}
\pagestyle{empty}

\begin{abstract}


Today's robot policies exhibit subpar performance when faced with the challenge of generalizing to novel environments.
Human corrective feedback is a crucial form of guidance to enable such generalization.
However, adapting to and learning from online human corrections is a non-trivial endeavor: not only do robots need to remember human feedback over time to retrieve the right information in new settings and reduce the intervention rate, but also they would need to be able to respond to feedback that can be arbitrary corrections about high-level human preferences to low-level adjustments to skill parameters. 
In this work, we present \emph{Distillation and Retrieval of Online Corrections} (\ACRO), a large language model (LLM)-based system that can respond to arbitrary forms of language feedback, distill generalizable knowledge from corrections, and retrieve relevant past experiences based on textual and visual similarity for improving performance in novel settings.
\ACRO is able to respond to a sequence of online language corrections that address failures in both high-level task plans and low-level skill primitives. 
We demonstrate that \ACRO effectively distills the relevant information from the sequence of online corrections in a knowledge base and retrieves that knowledge in settings with new task or object instances. \ACRO outperforms other techniques that directly generate robot code via LLMs~\cite{codeaspolicies2022} by using only half of the total number of corrections needed in the first round and requires little to no corrections after two iterations. We show further results and videos on our project website: \url{https://sites.google.com/stanford.edu/droc}.

\end{abstract}

\section{Introduction}
\label{sec:introduction}

From generating high-level plans to writing robot code -- pre-trained large language models (LLMs) have shown to exhibit a wide range of capabilities on robots that can in-context adapt to feedback and adjust to language corrections.
For example, InnerMonologue \cite{huang2022inner} and ReAct \cite{yao2022react} show that LLMs can use language feedback to modify task-level step-by-step plans, while Language-to-Reward 
demonstrates that LLMs can respond to low-level feedback by changing the reward function~\cite{yu2023languagetorewards}.
%
In many cases, these feedback mechanisms have shown to be important for improving LLM-driven policies in their capacity to adapt to new settings~\cite{tellex2020robots,sharma2022correcting,ren2023robots,kwon2023toward}.
%

\begin{figure*}[t]
    \centering
    \includegraphics[width=0.98\linewidth]{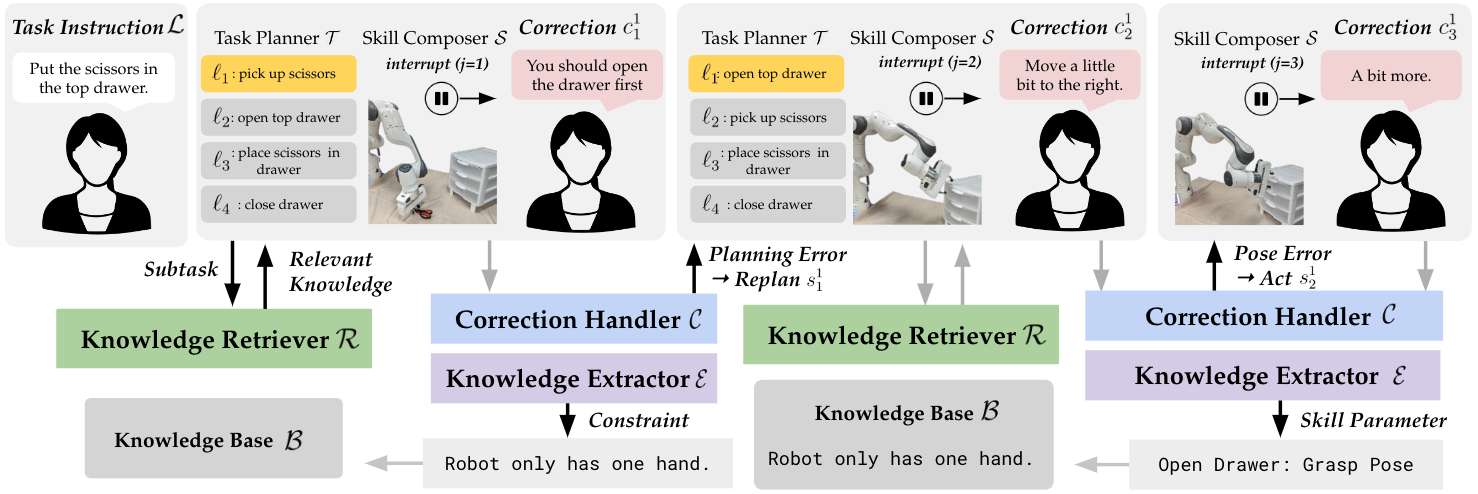}
    \caption{\footnotesize \textbf{Overview of \ACRO with example task ``put the scissors in the top drawer''}: the human interrupted the robot when it attempts to pick up the scissors before opening the drawer, the correction handler regenerated a plan accordingly and the knowledge extractor extracts a high-level constraint; during executing the skill of \textit{opening top drawer}, the human interrupted again to correct the grasping pose of the robot by providing two low-level commands.  }
    \label{fig:overview}
    \vspace{-0.4cm}
\end{figure*}

While prior work can respond to language corrections, they often take a rigid approach that does not allow for \textit{arbitrary forms} of human feedback. 
For instance, a human user might say ``Put the scissors in the top drawer'' as shown in \cref{fig:overview}, which leads to a robot planning on ``picking up the scissors''. However, if the drawer is not already open, the correct plan requires the robot to first open the drawer before picking up the scissors. A human observing the robot might decide to provide corrective language that addresses this planning error. With such a correction, the robot can finally proceed with the correct high-level plan, and attempt to ``open the top drawer''. However, the primitive that the robot executes might miss the grasp point on the handle. A human observer again can provide language corrections to guide the robot to finally achieve the skill primitive. 
Considering this example, it is non-trivial to interpret these different types of feedback: to interpret the first correction (\small{``You should open the drawer first''}\normalsize), one needs to know the robot's original plan; responding to the second correction (\small{``Move a little bit to the right''}\normalsize) requires the robot to infer what reference frame the human is using for directions; and for the next correction (\small{``a bit more''}\normalsize), the robot needs to know the history of action it took. 
The ability to respond to these arbitrary corrections requires a method that is flexible enough to identify if the corrections are about the high-level task or low-level skill primitives, and is able to leverage the prior context when responding to a sequence of online corrections.
In addition, corrections are predominantly handled in a \textit{temporary} fashion in prior work -- limited by what can fit within context length of the language model, and can be lost if not saved as the input context buffer overrides past feedback interactions with new ones.
%
As an example, in \cref{fig:overview}, the robot should be able to learn that it has only one arm and reuse this constraint when planning for future tasks.  
However, remembering the relevant information from a sequence of corrections can be challenging in more general settings beyond simple constraints. 
In this work, we address these challenges and enable LLM-based robot policies to respond to \textit{arbitrary} forms of feedback and further \textit{remember} and \textit{retrieve} feedback for future tasks. 
We present \ACRO, Distillation and Retrieval of Online Corrections
, a simple yet surprisingly effective formulation for responding to, remembering, and retrieving feedback. A key aspect of \ACRO is that it effectively uses an LLM to directly infer \emph{how-to-respond}, \emph{what-to-remember}, and \emph{what-to-retrieve}.
Specifically, \ACRO prompts an LLM to reason about relevant context to respond to online corrections,
distill language feedback into reusable knowledge 
and leverage visual similarities of objects for knowledge retrieval.
Experiments across multiple long-horizon manipulation tasks 
show that \ACRO excels at (i) responding to online corrections, and (ii) adapting to new objects and configurations while consistently reducing the number of human corrections needed over time. 
\ACRO also achieves higher success rates and requires fewer corrections in new tasks compared to baselines such as Code as Policies (CaP)~\cite{codeaspolicies2022} and variants with simpler implementation of distillation or retrieval mechanisms.
%
%
%

\section{Related Work}
\label{sec:related-work}


\noindent \textbf{LLM-powered Robotics.} 
Recent research has demonstrated planning capabilities of LLMs in robotics, including zero-shot generation of high-level task plans~\cite{huang2022language,wu2023tidybot,saycan2022arxiv,lin2023text2motion}, adapting based on human feedback when uncertain about the task plan~\cite{huang2022inner,ren2023robots}, and directly producing robot code~\cite{codeaspolicies2022,singh2022progprompt}. 
In addition, LLMs are used in a variety of settings beyond task planning in robotics and human-robot interaction such as reward design~\cite{kwon2023reward,hu2023language,yu2023languagetorewards}, in-context learning~\cite{mirchandani2023large}, and reasoning about multimodal inputs such as gestures~\cite{lin2023gesture}. 
While most prior works focus on leveraging LLMs for task-level reasoning or generating high-level primitives in code, recent works have also studied effectively leveraging LLMs along with vision language models (VLMs) for responding to fine-grained language such as ``move a bit to the right'' by either leveraging reward functions~\cite{yu2023languagetorewards}, voxelized representations~\cite{huang2023voxposer,shridhar2022peract}, or keypoints~\cite{sundaresan2023kite}.
On the other hand, a number of recent approaches have been leveraging VLMs directly as a success detector to ground the LLM plans and identify failures~\cite{liu2023reflect,guo2023doremi}. However, existing VLMs are not trained on manipulation data, and hence are not able to detect or provide fine-grained feedback to fix low-level mistakes such as missing a grasping point by a few centimeters. 
While some prior works address correcting low-level failures and others address planning-level errors, none of the prior work can tackle corrections at both levels. Meanwhile, prior literature does not consider knowledge distillation and long-term generalization from corrections for LLM-based robot policies, and is only concerned about immediate responses to language corrections.

\noindent \textbf{Knowledge Base for LLMs.}
Knowledge bases have previously been shown to be effective for retrieving examples when few-shot prompting LLMs~\cite{radford2019language,brown2020language,izacard2022few,wang2022language,zeng2022socratic}. Given a new example, in-context learning relies on the new example's similarity to previous examples to generate a response~\cite{chan2022transformers}.
This makes retrieval-augmentation rather straightforward for traditional Q\&A tasks – relevant examples can be sampled from the knowledge base by simply measuring the similarity between input questions (\eg via sentence embeddings \cite{devlin2018bert,choi2021evaluation,lan2019albert}). 
However to synthesize large amounts of feedback for robotics tasks, similarity-based methods are not enough; the method must also be able to summarize feedback~\cite{wang2023voyager,zhao2023expel}.
The design of \ACRO combines LLM capabilities previously demonstrated independently:
(i) summarizing multiple rounds of feedback (a mechanism shared by \cite{yao2022react,zhao2023expel}),
and (ii) to autonomously partition feedback into high-level or low-level to cover a broader range of re-usable adjustments~\cite{wang2023voyager} – in ways that together enable new modes of generalization from language feedback to new robot tasks or environments.

\noindent \textbf{Learning from Language Corrections.}
Literature outside LLM-based frameworks has also explored how to incorporate language corrections for adapting policies. 
A body of work developed methods for training a policy conditioned on past rollouts and language corrections so that it can iteratively improve using human feedback~\cite{co2018guiding,bucker2022reshaping,bucker2022latte} .
Prior methods also propose to learn cost maps for modifying robot plans with natural language~\cite{sharma2022correcting}.
Both of these categories of work learn how to update a policy or a plan through post-hoc offline language corrections. 
Our method responds to human online corrections that modify the robot's behavior as it executes, so that a user can help the robot to recover from mistakes.
However, there are a number of recent techniques that enable responding to real-time language corrections~\cite{broad2016towards, cui2023no,lynch2023interactive}. These works either make restrictive assumptions such as using distributed correspondence graphs to ground language, or they require extensive amount of language corrections and demonstrations.
In contrast, our work does not require training data for interpreting corrections, and leverages LLMs to directly generate robot code given the language corrections.
\section{\ACRO: \\ Distillation and Retrieval of Online Corrections}
\label{sec:method}
In this section, we first present our problem formulation, then present our method, \ACRO, by discussing how to generate robot plans and skills in response to language corrections, and describing how to distill and retrieve generalizable knowledge. 

\noindent \textbf{Problem Formulation.}
Consider a manipulation problem defined by a natural language instruction $\mathcal{L}$ (e.g., "put the scissors in the drawer"). 
Directly generating control sequences in response to this instruction can be extremely hard because $\mathcal{L}$ may be complex and long-horizon. 
We leverage a task planner $\mathcal{T}: \mathcal{L}\mapsto(\ell_1, \ell_2, \dots, \ell_M)$ to decompose the high-level instruction $\mathcal{L}$ into low-level skills $\ell_i$ (e.g., "open the drawer", "pick up the scissors"). For each skill, a skill composer maps the skill to the control policy $\mathcal{S}: \ell \mapsto p$. We follow Code-as-Policies (CaP) \cite{codeaspolicies2022} and represent $p$ as a code snippet. 
In practice, both $\mathcal{T}$ and $\mathcal{S}$ are not perfect and can make errors due to a variety of factors such as perception or control errors. 
A human interacting with this robot would only identify these errors \textit{while} the robot is executing the skill. As shown in~\cref{fig:overview}, the human only realizes the planner made a mistake when they see the robot attempt to pick up the scissors first. 
Once they spot the errors, the human can issue an arbitrary number of corrections during the execution of each skill $p_i$ until $\ell_i$ is fulfilled or the plan $P=(\ell_1, \ell_2, \dots, \ell_M)$ is correctly modified.
We use the subscript $j$ to stand for the round of corrections within the execution of $l_i$, and denote $c^i_j$ as the human language correction and $s^i_j$ as the solution to the correction. $s^i_j$ takes the form of two types of language programs: 1) triggering $\mathcal{T}$ to generate the correct plan, or 2) triggering $\mathcal{S}$ to execute another language skill.
At the end of correction round $j$, we define the interaction history as $H^i_j = \bigcup_{t=1:j}(P, \ell_i, p_i, c^i_t, s^i_t)$. 
We denote the total number of corrections at the end of the task to be $J$.
The goal of a learning agent is to reduce the amortized number of corrections across tasks:
    $\bar{J} = \frac{1}{N}\sum_{k=1}^{N} J_k,$
where $J_k$ is the total number of corrections at the end of task $\mathcal{L}_k \in \{\mathcal{L}_{1:N}\}$. 

\smallskip
\noindent \textbf{The \ACRO framework.}
\ACRO can be decomposed into three reasoning modules:  correction handler $\mathcal{C}$ (\textit{how-to-respond}), knowledge extractor $\mathcal{E}$ (\textit{what-to-remember}), and knowledge retriever $\mathcal{R}$ (\textit{what-to-retrieve}).
To generate the solution $s^i_j$ to the human correction $c^i_j$, we first extract relevant knowledge from the history $H^i_{j-1}$ with the correction handler $\mathcal{C}$ and decide whether it should generate a modified plan (triggering $\mathcal{T}$ with added constraint) or code policy (triggering $\mathcal{S}$ with relevant history) to address the user's correction, i.e., $s^i_j = \mathcal{C}(H^i_{j-1})$. 
Upon the completion of a plan or a skill, $\mathcal{E}$ distills generalizable knowledge from the full interaction history $H$ and saves it to the knowledge base $\mathcal{B}$. The saved knowledge can be retrieved later from $\mathcal{B}$ by the knowledge retriever $\mathcal{R}$ to guide task planning or skill composing.
We next discuss our task planner $\mathcal{T}$ and skill composer $\mathcal{S}$.

\smallskip
\noindent \textbf{Task planning with $\mathcal{T}$.}
To ground the output plan $P=(\ell_1, \ell_2, \dots, \ell_M)$, we provide scene information, few-shot examples, and rule constraints to guide the plan generation. Below is the template of the prompt\footnote[1]{Full prompts are available on our website \url{https://sites.google.com/stanford.edu/droc}.} along with an example:
\smallskip
\color{gray}
\begin{minted}[
% frame=single,
framesep=2mm,
baselinestretch=1.2,
fontsize=\scriptsize,
breaklines,
breakindentnchars=-1,
breaksymbolleft= ,
escapeinside=||,
% highlightlines={8-17},
% fontfamily=courier
]
{evoque}
Your role is to break down instructions into smaller sub-tasks.
# Examples: ...
Constraints:
1. The robot should manipulate one object and only move its gripper once in each sub-task.
2. If the instruction is ambiguous, first refer to the constraints to see whether you can replace the ambiguous reference; if not just leave it as is.
3. Tablewares should be put in the top drawer.
Object state: top drawer(closed), bottom drawer(open), spoon(on table), salt(in bottom drawer)
Instruction: put the spoon into the drawer
Plan:     |\textcolor{Green}{1: "Open the top drawer",}|
          |\textcolor{Green}{2: "Pick up the spoon",}|
          |\textcolor{Green}{3: "Put down the spoon into the top drawer",}|
          |\textcolor{Green}{4: "Close the top drawer"}|
\end{minted}
\color{black}
\smallskip
We define some initial constraints (1 and 2) in the prompt to enforce the hierarchical abstraction between $\mathcal{T}$ and $\mathcal{S}$, and handle human's ambiguous object reference. The initial object states are set to be "N/A", and we ask LLM to update the object states after each task execution or correction in the knowledge distillation phase (e.g., the object state becomes "top drawer(open)" after the successful execution of the task "open the drawer"). The constraints are also modifiable during that phase (e.g., constraint 3). In practice, one could use vision-language models (VLMs) as scene descriptors to obtain object states; however, we found existing open-source VLMs are sensitive to viewpoints and object locations, and require extensive prompt tuning. 

\smallskip
\noindent \textbf{Skill composing with $\mathcal{S}$.}
To ground the code policy $p_i$ generated by $\mathcal{S}$, we provide function APIs and few-shot examples to an LLM that generates the code corresponding to each skill primitive, similar to the practice in prior work~\cite{codeaspolicies2022}. 
The skill composer can use perception APIs to call a VLM-based perceiver to detect the task-related object.
For implementing the VLM perceiver, we use Segment-Anything ~\cite{kirillov2023segment} to obtain all objects' masks in the scene, and use CLIP~\cite{radford2021learning} to label each mask. 
\ACRO also provides task-related knowledge to $\mathcal{S}$, which can be used as primitive parameters to guide code generation. Such task-related knowledge is obtained from the knowledge distillation and retrieval phase, which will be discussed later.

\smallskip
\noindent \textbf{Correction handling with $\mathcal{C}$.}
Given a language correction $c^i_j$, \ACRO prompts the LLM to decide whether the correction is high-level (pertains to a constraint \eg ``robot can only grasp one object at a time'', user preference) or low-level (primitive parameter \eg relative gripper pose, or object information). 
If it's high-level, we pass the correction, the current skill and the original plan to the LLM for subsequent task planning. Otherwise if the correction is low-level, we ask the knowledge extractor to first extract relevant context from the (short-term) interaction history and then pass it as additional context to the LLM for subsequent code-writing.
A na\"ive instantiation of the knowledge extractor $\mathcal{E}$ would be 
\begin{figure}
    \centering
    \includegraphics[width=\linewidth]{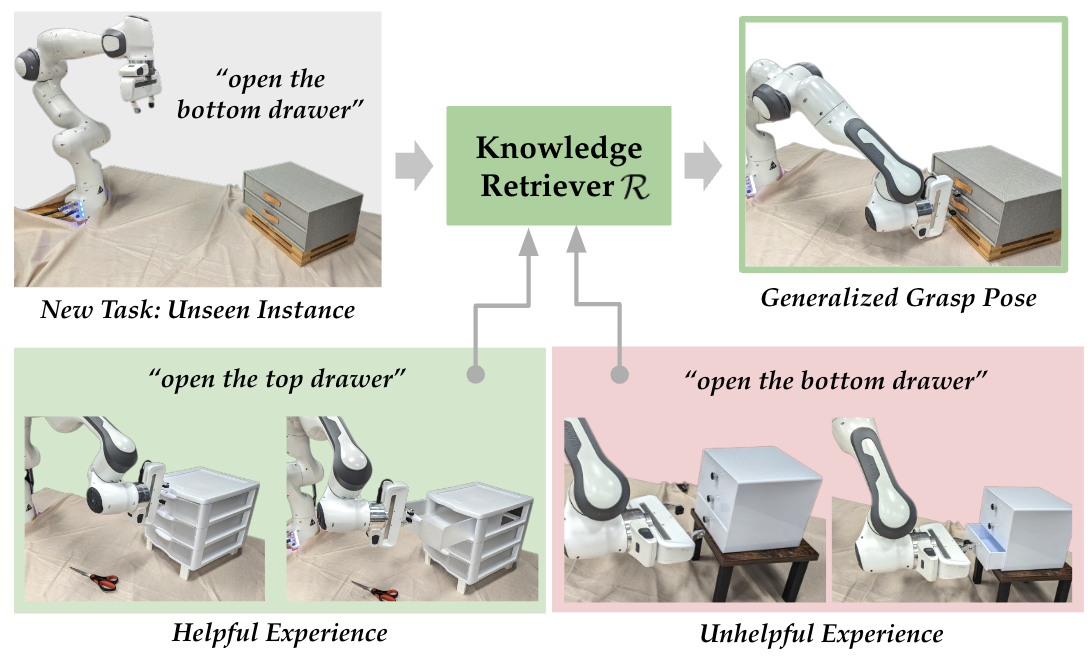}
    \caption{\textbf{Example of Visual Retrieval.} To retrieve the relevant task for opening the bottom gray drawer, textual similarity of the task instructions alone cannot filter the correct experience to reuse and similarity between visual features of the object (drawer handles specifically) are important for retrieving the correction past experience. }
    \label{fig:visual-retrieval}
    \vspace{-0.5cm}
\end{figure}
one that is prompted with the entire interaction history $H$, which results in lengthy context that buries important information required for reasoning.
At the same time we observe that the low-level correction $c^i_j$ alone is semantically meaningful enough for deciding what history to retrieve for interpreting it. Thus, we only provide $c^i_j$ to the LLM, and limit the retrievable history to four categories: (a) Last interaction. (b) Initial interaction. (c) No dependence.
\smallskip
\color{gray}
\begin{minted}[
% frame=single,
framesep=2mm,
baselinestretch=1.2,
fontsize=\scriptsize,
breaklines,
breakindentnchars=-1,
breaksymbolleft= ,
escapeinside=||,
% highlightlines={9-15},
% fontfamily=courier
]
{evoque}
A human is issuing corrections to a robot, which encounters errors during executing a task. These corrections may depend on the robot's past experiences. Your task is to determine what does a correction depend on: (a) Last interaction. (b) Initial interaction. (c) No dependence. 
# Examples: ...
"Move right a little bit": |\textcolor{Green}{(c)}|
"Keep going": |\textcolor{Green}{(a)}|
"Now you can continue": |\textcolor{Green}{(b)}|
\end{minted}
\color{black}
\smallskip

Once $H^i_r$ is retrieved, $\mathcal{C}$ will prompt the LLM again to generate the solution $s^i_j$.
We implement two types of correction handlers $\mathcal{C}_\mathcal{T}$ and $\mathcal{C}_\mathcal{S}$ depending on the error level, and $s^i_j$ can be a re-plan or a new code policy. 
The prompt structures of $\mathcal{C}_\mathcal{T}$ and $\mathcal{C}_\mathcal{S}$ are similar to that of $\mathcal{T}$ and $\mathcal{S}$, with ($H^i_r$, $c^i_j$) and additional guidance concatenated at the end of the prompts.
Once $s^i_j$ is generated, we execute it on the robot, and repeat the above process until either the skill $\ell_i$ is fulfilled or the plan $P$ has been successfully modified as deemed by the human.

To enable easy corrections and adaptations, we ground each correction to the related object or the task. For fine-grained corrections, we prompt $\mathcal{C}_\mathcal{S}$ to reason about whether the reference frame is an absolute frame or an object-centric frame. We empirically observe that human users tend to issue corrections in the object-centric frame. For example, when the robot is executing the task ``open the drawer'', the correction ``move forward a bit'' should be interpreted as ``move towards the drawer a bit'', which requires the knowledge of the drawer's frame to properly deal with the correction. 
To obtain the object-centric frame, we provide the LLM with the object's geometric properties (e.g., "normal vector") and ask it to represent the object's frame with these properties. We also ground the scale of movements to the size of the related object whenever the correction contains vague distance expressions (e.g., ``a little bit''), which makes our system more adaptable.


\smallskip

\noindent \textbf{Knowledge Distillation with $\mathcal{E}$.} 
Given a history log with task instructions, generated LLM outputs, and several rounds of feedback interleaved, \ACRO prompts the LLM to summarize task-related knowledge, variables to save, modified code/plan, and updated states, then stores this relevant information into the knowledge base.
At the end of each skill or plan, we provide the skill description $\ell_i$ and the interaction history $H$ to $\mathcal{E}$ and use chain-of-thought prompting~\cite{wei2023chainofthought} to first reason about what are the types of knowledge that can be generalized to similar tasks in the future, and then extract the values of these types of knowledge from $H$. Below is the template of the prompt:
\smallskip
\color{gray}
\begin{minted}[
% frame=single,
framesep=2mm,
baselinestretch=1.2,
fontsize=\scriptsize,
breaklines,
breakindentnchars=-1,
breaksymbolleft= ,
escapeinside=||,
% highlightlines={9-15},
% fontfamily=courier
]
{evoque}
Your task is to extract reusable knowledge from the provided interaction history.
# Examples: ...
Task name: {TASK_NAME}
Task-related knowledge: |\textcolor{Green}{# LLM's answer here}|
Interaction history: {HISTORY}
Variables to save: |\textcolor{Green}{# LLM's answer here}|
Modified code/plan: |\textcolor{Green}{# LLM's answer here}|
Updated object state: |\textcolor{Green}{# LLM's answer here}|
\end{minted}
\color{black}
\smallskip

In our implementation, we separate the plan-level distillation and the skill-level distillation prompts. Examples of plan-level knowledge include task constraints (e.g., ``the scissors cannot be put in a full drawer''), robot constraints (e.g., ``the robot can only grasp one thing at a time''), and user preferences (e.g., ``The user prefer milk to coke''). Examples of skill-level knowledge are task parameters (e.g., gripper pose, pull distance) and object information  (e.g., visual feature, label). The distilled knowledge is saved to the knowledge base $\mathcal{B}$ in a dictionary format, with the key to be each task's name. We also ask $\mathcal{E}$ to update the objects' states after each skill is successfully performed. For ambiguous references, we remember the label of the object and also treat this information as a task constraint.

\smallskip

\noindent \textbf{Knowledge Retrieving with $\mathcal{R}$.}
Given a new task, \ACRO prompts the LLM to decide which past experiences are relevant. In addition, our knowledge retriever leverages visual similarities for measuring relevancy when language alone is not sufficient.
When queried by an instruction or a skill description, the knowledge retriever $\mathcal{R}$ indexes into the knowledge base $\mathcal{B}$ and retrieves relevant knowledge to help address the query. There are two retrieval metrics: (a) task semantics; (b) visual feature of the task-related object. Some knowledge is shared across different tasks (e.g., robot constraints, user preferences) and can always be retrieved, while other types of knowledge are specific to individual task category or object. The intuition here is that only knowledge that is both semantically and visually similar to the query can be retrieved. To implement this intuition, we first prompt $\mathcal{R}$ in a zero-shot manner to pick all the tasks that are semantically similar to the query:
\smallskip
\color{gray}
\begin{minted}[
% frame=single,
framesep=2mm,
baselinestretch=1.2,
fontsize=\scriptsize,
breaklines,
breakindentnchars=-1,
breaksymbolleft= ,
escapeinside=||,
% highlightlines={9-12},
% fontfamily=courier
]
{evoque}
I'll give you a list of tasks a robot has previously performed and a new task to address. Your goal is to determine the following:
1. Does the new task fall into the same category with any previous task? (E.g. "open" and "put" are different categories of tasks)
2. If so, which specific previous tasks are they? Answer in list format.
Previous tasks: 1. Open the top drawer. 2. Pick up the scissors. 3. Put the mug on the shelf. 4. Pick up the yellow marker.
New task: Pick up the spoon.
Response:
    |\textcolor{Green}{1: "Yes",}| |\textcolor{Green}{2: [2, 4]}|
\end{minted}
\color{black}

Then, we can compare the queried object's visual feature to visual features of objects from the chosen tasks, and retrieve the one with highest visual similarity to add to the prompt of $\mathcal{T}$ or $\mathcal{S}$ as guidance. 
We motivate our design choice of using visual features for retrieval with the example shown in \cref{fig:visual-retrieval}. In order to ``open the bottom drawer'' shown in the image, the robot needs to retrieve a grasp pose that associates with a horizontal drawer handle instead of a knob. It cannot retrieve the correct information using only skill or task similarity. 
Another advantage of visual retrieval is when the queried object's name contains visual information (e.g., "white drawer"), we can calculate the semantic-visual similarity between the name and the visual features of the chosen tasks to choose the most similar one. 
It is also important to note that different pre-trained vision or vision-language models encode different features. For manipulation tasks, we often care about the state and shape of an object more than the texture similarity. We use DINO-V2~\cite{oquab2023dinov2} features out of the box for visual-visual retrieval, and use CLIP features for visual-semantic retrieval.
\section{Experiments}
\label{sec:experiments}


\begin{figure*}
\centering
    \includegraphics[width=0.93\linewidth]{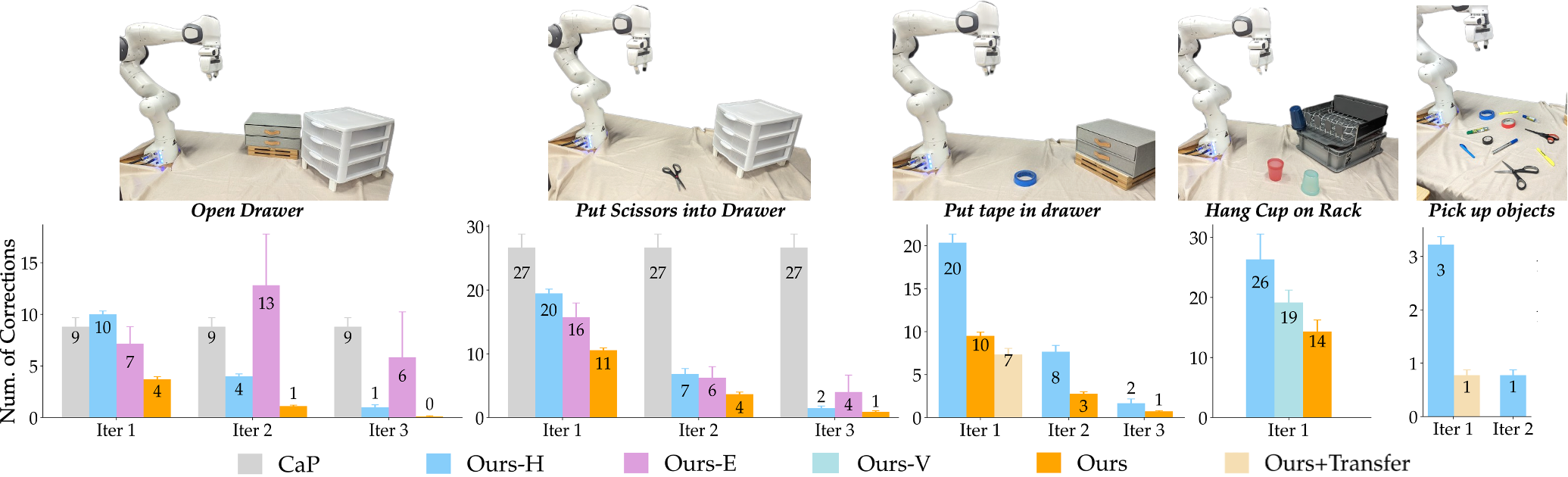}
    \caption{\footnotesize \textbf{Skill-level results.} For all tasks, the results are averaged over six rounds of experiments. The error bars reflect the standard errors across different rounds. Each iteration corresponds to a different task setting. The number of corrections declines as the iteration increases, which shows that \textit{\textbf{\ACRO}} can generalize and adapt to unseen new settings. For the ``Hang Cup on Rack'' task, we are not showing decline of corrections over iterations but instead ablate the correction and distillation module of our system.}
    \label{fig:skill-level}
    \vspace{-0.2cm}
\end{figure*}

\begin{figure*}
\centering
    \includegraphics[width=0.93\linewidth]{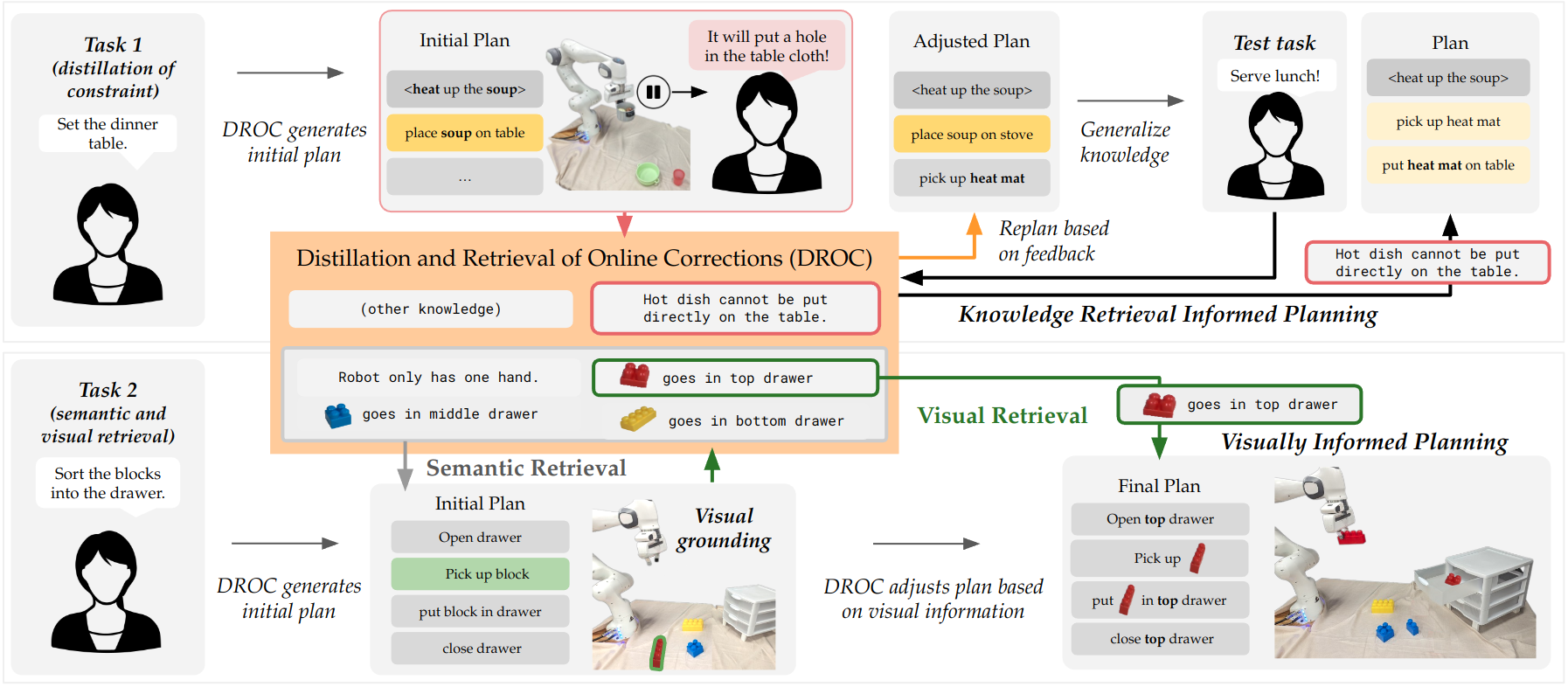}
    \caption{\footnotesize \textbf{Illustrative examples for plan-level test cases.} (1) upon interruption, \textit{\textbf{\ACRO}} responds to correction by identifying it is a plan-level error and replanning, and distills the constraint for future tasks; (2) given a task with ambiguity, \textit{\textbf{\ACRO}} retrieves past experiences base on semantic and visual similarities.}
    \label{fig:plan-level}
    \vspace{-0.4cm}
\end{figure*}

We evaluate our approach on a real-world tabletop environment with a Franka Emika Panda Robot. We use GPT-4~\cite{openai2023gpt4} for all LLM modules. We design experiments to test \textit{\textbf{\ACRO}}'s core capabilities: 1) accurately responding to online corrections, 2) distilling generalizable knowledge, and 3) retrieving relevant knowledge in novel tasks to improve performance. 


\noindent \textbf{Tasks.} 
We will test \textit{\textbf{\ACRO}}'s ability to respond to both skill-level and plan-level corrections using separate set of tasks summarized in \cref{tab:skill-level-tasks}.
For skill-level, we experiment with 5 table-top manipulation tasks, as shown in top row of \cref{fig:skill-level}.

We iterate each task 3 times. We start with an initial setting and issue corrections until the first iteration is fulfilled. Then, with the distilled knowledge from last iteration, we change the setup (objects' location, instance etc.) and repeat the task.
%

\begin{table}[]
\footnotesize
\caption{Summary of skill-level and plan-level tasks.}
\renewcommand{\arraystretch}{1.08}
\resizebox{\linewidth}{!}{
\begin{tabular}{ccc}
\hline
\textbf{Skill-level Tasks}    & \textbf{Object Variations}   & \textbf{Knowledge}          \\ \hline

 \multicolumn{1}{c}{Open drawer}                                                           & \multicolumn{1}{c}{2 drawers}  & \multicolumn{1}{c}{{Grasp, Pull}}                                               \\

 \multicolumn{1}{c}{Put scissors in drawer}                                                  & \multicolumn{1}{c}{2 scissors, 2  drawers}  & \multicolumn{1}{c}{{Grasp, Pull, Place}}                                                \\
 \multicolumn{1}{c}{Put tape in drawer}                                                  & \multicolumn{1}{c}{5 tapes, 2 drawers}    & \multicolumn{1}{c}{{Grasp, Pull, Place}}                                                   \\
 \multicolumn{1}{c}{Hang cup on the rack}                                                  & \multicolumn{1}{c}{Flipped cup, upright cup}  &\multicolumn{1}{c}{{Grasp, Place}}                                  \\

 \multicolumn{1}{c}{Pick up object}                                                        & \multicolumn{1}{c}{6 objects}      & \multicolumn{1}{c}{{Grasp}}               \\
 
\vspace{-0.15cm} \\
\hline

 \multicolumn{1}{c}{\textbf{Plan-level Tasks}} & \multicolumn{1}{c}{\textbf{Test Tasks}} & \multicolumn{1}{c}{\textbf{Knowledge}}  \\ \hline

\multicolumn{1}{c}{Put scissors in drawer}          & \multicolumn{1}{c}{Clean the table}                     & \multicolumn{1}{c}{\multirow{2}{*}{Pref.}}                      \\
\multicolumn{2}{c}{\textcolor{orange}{\textit{"User wants stationery in white drawer"}}}                                         & \multicolumn{1}{c}{} \vspace{0.13cm}\\

\multicolumn{1}{c}{Bring cup of coffee}          & \multicolumn{1}{c}{Make cup of coffee}                     & \multicolumn{1}{c}{\multirow{2}{*}{Pref. + Feasi.}}                      \\
\multicolumn{2}{c}{\textcolor{orange}{\textit{"User doesn't drink black coffee"}}}                                         & \multicolumn{1}{c}{} \vspace{0.13cm}\\

\multicolumn{1}{c}{Heat milk in fridge}          & \multicolumn{1}{c}{Slice carrot}                     & \multicolumn{1}{c}{\multirow{2}{*}{Feasi.}}                      \\
\multicolumn{2}{c}{\textcolor{orange}{\textit{"Robot only has one hand"}}}                                         & \multicolumn{1}{c}{} \vspace{0.13cm}\\

\multicolumn{1}{c}{Sort blocks to drawer}          & \multicolumn{1}{c}{Sort blocks to drawer}                     & \multicolumn{1}{c}{\multirow{2}{*}{Feasi.}}                      \\
\multicolumn{2}{c}{\textcolor{orange}{\textit{"Same color block goes to same drawer"}}}                                         & \multicolumn{1}{c}{} \vspace{0.13cm}\\

\multicolumn{1}{c}{Put shoes on rack}          & \multicolumn{1}{c}{Sort clothes into shelf}                     & \multicolumn{1}{c}{\multirow{2}{*}{Comm.}}                      \\
\multicolumn{2}{c}{\textcolor{orange}{\textit{"Same types of clothing go to same place"}}}                                         & \multicolumn{1}{c}{} \vspace{0.13cm}\\

\multicolumn{1}{c}{Set dinner table}          & \multicolumn{1}{c}{I want to have lunch}                     & \multicolumn{1}{c}{\multirow{2}{*}{Comm. + Scene. + Feasi.}}                      \\
\multicolumn{2}{c}{\textcolor{orange}{\textit{"Fork on left, hot dish on heat mat"}}}                                         & \multicolumn{1}{c}{} \vspace{0.13cm}\\

\multicolumn{1}{c}{Place book on shelf}          & \multicolumn{1}{c}{Put DVD on shelf}                     & \multicolumn{1}{c}{\multirow{2}{*}{Scene.}}                      \\
\multicolumn{2}{c}{\textcolor{orange}{\textit{"White shelf is full"}}}                                         & \multicolumn{1}{c}{} \vspace{0.06cm}\\

\hline
\end{tabular}
}
\label{tab:skill-level-tasks}
\vspace{-0.5cm}
\end{table}

To evaluate \textit{\textbf{\ACRO}} on plan-level corrections, we design seven long-horizon tasks (see \cref{fig:plan-level} for examples) to test four types of generalization of knowledge: (a) \textit{User preferences}, which we aims to investigate whether \textit{\textbf{\ACRO}} can distill user-specific knowledge and apply it to future tasks; (b) \textit{Feasibility of plans}, which we want to see whether \textit{\textbf{\ACRO}} can understand the constraints of different tasks and the robot from corrections; (c) \textit{Common-sense reasoning}, which we aim to test if \textit{\textbf{\ACRO}} can ground LLMs' powerful common-sense knowledge to robotics tasks by explicitly distilling task-related common-sense knowledge from corrections; (d) \textit{Scene information}, which we aims to see if \textit{\textbf{\ACRO}} can understand the scene by distilling scene information from corrections. Orange texts in \cref{tab:skill-level-tasks} show the knowledge distilled from train tasks (i.e., tasks on the left) that can be retrieved for test tasks.


\noindent \textbf{Baselines.} For skill-level tasks, we compare with the following baselines: (a) \textit{\textbf{CaP}}: Code as Policies \cite{codeaspolicies2022}, 
which handles corrections as new instructions and does not have a knowledge distillation or retrieval module; 
(b) \textit{\textbf{Ours--H}}: \textit{\textbf{\ACRO}} with no initial history to show that knowledge distillation from prior tasks are important; (c) \textit{\textbf{Ours--E}}: \textit{\textbf{\ACRO}} without relevant context extractor (uses all interaction history for correction handling), and (d)  \textit{\textbf{Ours--V}}: \textit{\textbf{\ACRO}} that does not leverage visual retrieval. 
The baselines share the same prompts with our task planner $\mathcal{T}$ and skill composer $\mathcal{S}$, and have access to exactly the same function APIs. 
For plan-level tasks, we compare with \textit{\textbf{Ours--R}}, an ablation that does not distill knowledge and naively retrieves saved plans. 

\noindent \textbf{Skill-Level Corrections.} 
%
We report the amortized number of corrections for each task over learning iterations in \cref{fig:skill-level}. 
Three of the authors served as oracle users to provide feedback. The same user provided feedback within the same task such that the corrections are consistent across iterations.
Overall, \textit{\textbf{\ACRO}} outperforms all baselines by requiring less number of corrections, which strongly supports that \textit{\textbf{\ACRO}} can synthesize generalizable knowledge from the corrections and use them to quickly adapt to unseen task settings. 
We further analyze the results for evaluating the core capabilities of \textit{\textbf{\ACRO}} and our specific design choices:

\noindent \textit{\textbf{\textit{\textbf{\ACRO}} enables more effective corrections.}} The comparison between \textit{\textbf{\ACRO}} (ours) and \textit{\textbf{CaP}} suggests that \textit{\textbf{\ACRO}} can respond to corrections significantly more effectively as it requires less than half corrections in the first round. 

\noindent \textit{\textbf{\textit{\textbf{\ACRO}} distills generalizable knowledge within the same task (same language, same object).}}
The comparison between the initial corrections needed for \textit{\textbf{\ACRO}} (ours) and \textit{\textbf{Ours--H}} demonstrates that the distilled knowledge helps with learning different variations of the same task.

\noindent \textit{\textbf{Visual retrieval enables more accurate skill generalization.}} For the "Hang cup on Rack" task, 
we provide the system with 2 sets of knowledge with different visual features (cup's color, cup's orientation). Because upright cups and flipped cups require different policies (the robot needs to flip the upright cup before hanging it), our system needs to retrieve the correct knowledge through visually comparing cups' orientation at the presence of distraction from cups' colors. 
Through ablating with \textit{\textbf{Ours--V}}, we show that visual retrieval is an important element in \textit{\textbf{\ACRO}}.

\noindent \textit{\textbf{\textit{\textbf{\ACRO}} distills generalizable skill parameters across objects.}} We further tested cross-object transfer in ``pick up object'' and ``put tape in drawer'' tasks and report results as \textit{\textbf{Ours+Transfer}}. Specifically, for pick up object, we reuse history from other objects and for ``put tape in drawer'' , we reuse history from ``put scissors in top white drawer''. By reusing history from different tasks, our method further reduces the number of online corrections needed.

\noindent \textbf{Plan-Level Corrections.} We evaluate how \textit{\textbf{\ACRO}} respond to plan-level corrections and generalize to new tasks with a suite of task scenarios we curated to cover a diverse range of potential mistakes. The train tasks and test tasks can be found in~\cref{tab:skill-level-tasks}. 
We repeat 3 trials for each task scenario that begins with the train task and then move to the test task with the knowledge distilled from the train task. We only report the average number of corrections needed for the test tasks in ~\cref{tab:plan-level} because \textit{\textbf{\ACRO}} and \textit{\textbf{Ours--R}} share the same correction module and perform exactly the same on train tasks. 
The comparison between \textit{\textbf{\ACRO}} and \textit{\textbf{Ours--R}} again shows that \textit{\textbf{\ACRO}} can distill and retrieve knowledge that empower generalization, leading to smaller number of corrections required on the test tasks. 
We don't ablate \ACRO's correction module here because without that it's too difficult for the task planner to understand humans' corrections so as to modify the original plans.
We visually illustrate two cases in~\cref{fig:plan-level}: one demonstrating the effectiveness of knowledge distillation and retrieval of high-level constraint and the other showcasing how semantic retrieval and visual retrieval aid high-level task planning.

\vspace{-0.3cm}
\begin{table}[htbp]
\centering 
\footnotesize
\caption{Number of corrections required for plan-level test tasks.}
\renewcommand{\arraystretch}{1.1}
\begin{tabular}{ccc}
\hline
      \textbf{Task Types}            & \textbf{Ours} & \textbf{Ours--R} \\ \hline
User Preferences       & \textbf{0.5}             &               1              \\
Feasibility of Plans       & \textbf{0.67}            &                   1.33         \\
Common-sense Reasoning      & \textbf{0}            &                    1.5         \\
Scene Information & \textbf{0.5}             &                  1.5         \\ \hline
\end{tabular}
\label{tab:plan-level}
    \vspace{-0.6cm}
\end{table}




\vspace{10pt}
\section{Conclusion}
\label{sec:discussion}

\noindent \textbf{Discussion.} We propose \ACRO, an LLM-based system for enabling robots to respond to arbitrary forms of online language corrections, distill generalizable information, and retrieve relevant knowledge for novel tasks. 
We demonstrate that \ACRO outperforms baselines for responding to both low-level and high-level corrections, and can effectively generalize knowledge within the same tasks and across similar tasks.

\smallskip

\noindent \textbf{Acknowledgement}
This project was sponsored by NSF Awards 1941722, 2132847, 2006388, ONR, AFOSR YIP, DARPA YFA, and Ford. 
Any opinions, findings,
conclusions or recommendations expressed in this material are those of the authors and do not necessarily reflect the views of the sponsors.

\small
\bibliographystyle{IEEEtranN}
\bibliography{reference}  


\end{document}